\pdfoutput=1

\documentclass[11pt]{article}

\usepackage{naacl2021}

\usepackage{times}
\usepackage{latexsym}

\usepackage[T1]{fontenc}

\usepackage[utf8]{inputenc}

\usepackage{microtype}

\usepackage{microtype}
\usepackage{booktabs}
\usepackage{subcaption}
\usepackage{multirow}
\usepackage{algorithm}
\usepackage{algpseudocode}
\usepackage{xcolor}
\usepackage{textcomp}
\usepackage{amsmath,amssymb,amsfonts}
\usepackage{rotating}
\usepackage{array}
\usepackage{makecell}
\usepackage{enumitem}
\usepackage{balance}
\newcommand{\subscript}[2]{$#1#2$}

\newcommand{\ie}{{\em i.e.}, }
\newcommand{\eg}{{\em e.g.}, }

\begin{document}
\title{\textsc{Edge}: Enriching Knowledge Graph Embeddings with External Text}

\author{Saed Rezayi$^1$, Handong Zhao$^2$, Sungchul Kim$^2$, Ryan A. Rossi$^2$, \\
  \textbf{Nedim Lipka}$^2$, and \textbf{Sheng Li}$^1$ \\
  $^1$Department of Computer Science, University of Georgia, Athens, GA, USA \\
  $^2$Adobe Research, San Jose, CA, USA \\
  \texttt{\{saedr,sheng.li\}@uga.edu} \\
  \texttt{\{hazhao,sukim,ryrossi,lipka\}@adobe.com} \\
}

\maketitle

\begin{abstract}
Knowledge graphs suffer from sparsity which degrades the quality of representations generated by various methods. While there is an abundance of textual information throughout the web and many existing knowledge bases, aligning information across these diverse data sources remains a challenge in the literature. Previous work has partially addressed this issue by enriching knowledge graph entities based on \textit{``hard''} co-occurrence of words present in the entities of the knowledge graphs and external text, while we achieve \textit{``soft''} augmentation by proposing a knowledge graph enrichment and embedding framework named \textsc{Edge}. Given an original knowledge graph, we first generate a rich but noisy augmented graph using external texts in semantic and structural level. To distill the relevant knowledge and suppress the introduced noise, we design a graph alignment term in a shared embedding space between the original and augmented graph. To enhance the embedding learning on the augmented graph, we further regularize the locality relationship of target entity based on negative sampling.
Experimental results on four benchmark datasets demonstrate the robustness and effectiveness of \textsc{Edge} in link prediction and node classification. 

\end{abstract}

\section{Introduction}
\label{sec:intro}
\begin{figure}[t]
    \centering
    \vspace{-2mm}
    \includegraphics[width=\columnwidth]{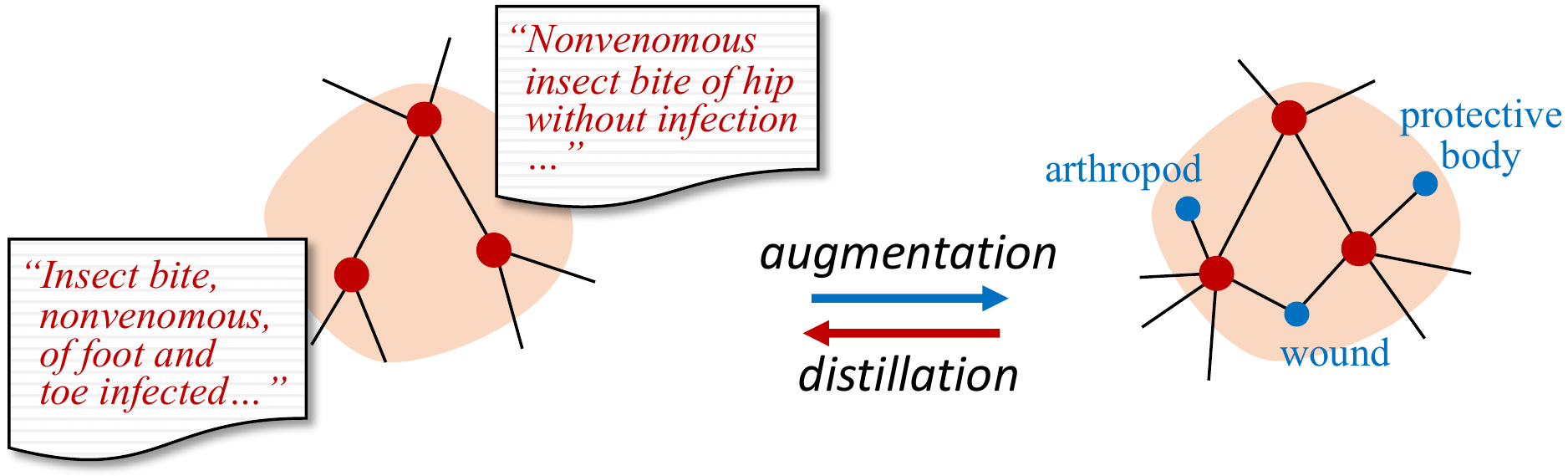}
    \vspace{-6mm}
    \caption{An example illustrating the original (left) and augmented knowledge graphs (right). Red nodes are knowledge graph entities and small blue nodes are textual nodes obtained from the external text. In augmentation process, a new set of keywords are discovered and attached to the original entities. To keep the augmented graph semantically close to the original graph, a backward pass of knowledge distillation is achieved by the proposed graph alignment.}
    \label{fig:gt}
    \vspace{-4mm}
\end{figure}

Knowledge Graph (KG)\footnote{Knowledge graph usually represents a heterogeneous multigraph whose nodes and relations can have different types. However in the work, we follow \cite{kartsaklis2018mapping}, consider knowledge graph enrichment problem where only one relation type (connected or not) appears.} embedding learning has been an emerging research topic in natural language processing, which aims to learn a low dimensional latent vector for every node.
One major challenge is sparsity. Knowledge graphs are often incomplete, and it is a challenge to generate low-dimensional representations from a graph with many missing edges. 
To mitigate this issue, auxiliary texts that are easily accessible have been popularly exploited for enhancing the KG (as illustrated in Figure \ref{fig:gt}). More specifically, given that KG entities contain textual features, we can link them to an auxiliary source of knowledge, \eg WordNet, and therefore enhance the existing feature space. With notable exceptions, the use of external textual properties for KG embedding has not been extensively explored before. Recently, \cite{kartsaklis2018mapping} used entities of the KG to query BabelNet \cite{Navigli2012babelnet}, added new nodes to the original KG based on co-occurrence of entities, and produced more meaningful embeddings using the enriched graph. However, this hard-coded, co-occurrence based KG enrichment strategy fails to make connections to other semantically related entities. As motivated in Figure \ref{fig:gt}, the newly added entities ``\textit{wound}", ``\textit{arthropod}" and ``\textit{protective body}", are semantically close to some input KG entity nodes (marked in red). However, they cannot be directly retrieved from BabelNet using co-occurrence matching.

In this paper, we aim to address the sparsity issue by integrating a learning component into the process. We propose a novel framework, \textsc{Edge}, for KG enrichment and embedding. \textsc{Edge} first constructs a graph using the external text based on similarity and aligns the enriched graph with the original KG in the same embedding space. It infuses learning in the knowledge distillation process by graph alignment, ensuring that similar entities remain close, and dissimilar entities get as far from each other. Consuming information from an auxiliary textual source helps improve the quality of final products, \ie low dimensional embeddings, by introducing new features. This new feature space is effective because it is obtained from a distinct knowledge source and established based on affinity captured by the learning component of our model.

More specifically, our framework takes $\mathcal{KG}$, and an external source of texts, $\mathcal{T}$, as inputs, and generates an augmented knowledge graph, $a\mathcal{KG}$. in generating $a\mathcal{KG}$ we are mindful of semantic and structural similarities among $\mathcal{KG}$ entities, and we make sure it contains all the original entities of $\mathcal{KG}$. This ensures that there are common nodes in two graphs which facilitates the alignment process. To align $\mathcal{KG}$ and $a\mathcal{KG}$ in the embedding space, a novel multi-criteria objective function is devised. In particular, we design a cost function that minimizes the distance between the embeddings of the two graphs. As a result, textual nodes (\eg blue nodes in Figure~\ref{fig:gt}) related to each target entity are rewarded while unrelated ones get penalized in a negative sampling setting.

Extensive experimental results on four benchmark datasets demonstrate that \textsc{Edge} outperforms state-of-the-art models in different tasks and scenarios, including link prediction and node classification. Evaluation results also confirm the generalizability of our model. We summarize our contributions as follows: (i) We propose \textsc{Edge}, a general framework to enrich knowledge graphs and node embeddings by exploiting auxiliary knowledge sources. (ii) We introduce a procedure to generate an augmented knowledge graph from external texts, which is linked with the original knowledge graph. (iii) We propose a novel knowledge graph embedding approach that optimizes a multi-criteria objective function in an end-to-end fashion and aligns two knowledge graphs in a joint embedding space. (iv) We demonstrate the effectiveness and generalizability of \textsc{Edge} by evaluating it on two tasks, namely link prediction and node classification, on four graph datasets.

The rest of the paper is organized as follows. In the next section, we try to identify the gap in the existing literature and motivate our work. Next, in Section \ref{sec:model}, we set up the problem definition and describe how we approach the problem by in-depth explanation of our model. We evaluate our proposed model by experimenting link prediction and node classification on four benchmark datasets and present the results and ablation study in Section \ref{sec:experiment}. Finally, we conclude our work and give the future direction in Section \ref{sec:con}.

\section{Related Work}
\label{sec:rw}

Knowledge graph embedding learning has been studied extensively in the literature~\cite{bordes2013transe,wang2014knowledge,yang2015embedding,sun2018rotate,zhang2019iteratively,xian2020cafe,yan2020graph,sheu2020context}. A large number of them deal with the heterogeneous knowledge graph, where it appears different types of edges. While in this work we consider the type of knowledge graph with only one type (i.e. connected or not) of relation, and only focus on entity embedding learning. Our work is related to graph neural networks, such as the graph convolutional networks (GCN)~\cite{kipf2017semi} and its variants~\cite{wu2020comprehensive,CensNetIJCAI,CensNetPAMI}, which learn node embeddings by feature propagation. In the following, we mainly review the most relevant works in two aspects, i.e., graph embedding learning with external text and knowledge graph construction.

\subsection{Graph Embedding with External Text}
The most similar line of work to ours is where an external textual source is considered to enrich the graph and learn low dimensional graph embeddings using the enriched version of the knowledge graph. For instance, \cite{wang2016text} annotates the KG entities in text, creates a network based on entity-word co-occurrences, and then learns the enhanced KG. Similarly, \cite{kartsaklis2018mapping} adds an edge $(e,t)$ to KG per entity $e$ based on co-occurrence and finds graph embeddings using random walks. However, there is no learning component in these approaches in constructing the new knowledge graph. And the enrichment procedure is solely based on occurrences (``hard" matching) of entities in the external text.

\begin{figure*}[t]
    \centering
    \includegraphics[width=0.9\textwidth]{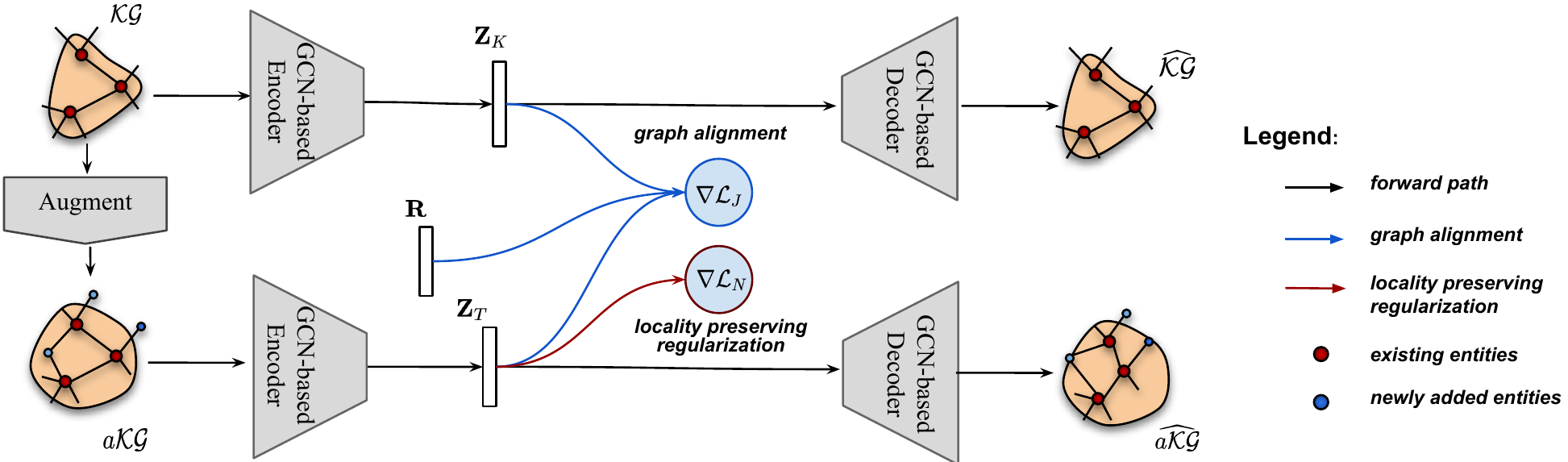}
    \vspace{0mm}
    \caption{Our proposed framework for aligning two graphs in the embedding space. The graph alignment component, $\mathcal{L}_J$, requires an additional matrix, $\mathbf{R}$, that selects embeddings of $\mathcal{KG}$ entities from $\mathbf{Z}_T$, so the resulting matrix, $\mathbf{R}\mathbf{Z}_T$, would have the same dimension as $\mathbf{Z}_K$. Furthermore, $\mathcal{L}_N$ penalizes additional entities that are unrelated to the target entity, while rewards the related ones. We omit the graph reconstruction loss for simplicity.}
    \label{fig:model}
    \vspace{-4mm}
\end{figure*}

For graph completion task, \cite{malaviya2020commonsense} uses pre-trained language models to improve the representations and for Question Answering task, \cite{sun2018open} extracts a sub-graph $\mathcal{G}_q$ from KG and Wikipedia, which contains the answer to the question with a high probability and apply GCN on $\mathcal{G}_q$ which is limited to a specific task. 
We emphasize that the main difference between our model and previous work is that we first create an augmented knowledge graph from an external source, and improve the quality of node representation by jointly mapping two graphs to an embedding space. To the best of our knowledge, this is the first time that a learning component is incorporated to enriching knowledge graphs. 

\subsection{Knowledge Graph Construction}
Knowledge graph construction methods are broadly classified into two main groups: 1) Curated approaches where facts are generated manually by experts, \eg WordNet \cite{Fellbaum1998} and UMLS \cite{bodenreider2004unified}, or volunteers such as Wikipedia, and 2) Automated approaches where facts are extracted from semi-structured text like DBpedia \cite{auer2007dbpedia}, or unstructured text \cite{carlson2010toward}. The latter approach can be defined as extracting structured information from unstructured text.
In this work, we do not intend to construct a knowledge base from scratch, instead we aim to generate an augmented knowledge graph using side information. Hence, we employ existing tools to acquire a set of new facts from external text and link them to an existing KG.

\section{Proposed Model}
\label{sec:model}
\subsection{Problem Statement}
We formulate the knowledge graph enrichment and embedding problem as follows: given a knowledge graph $\mathcal{KG}=(\mathcal{E},\mathcal{R},X)$ with $|\mathcal{E}|$ nodes (or entities), $|\mathcal{R}|$ edges (or relations) and $X\in \mathbb{R}^{|\mathcal{E}|\times D}$ as feature matrix, where $D$ is the number of features per entity, also given an external textual source, $\mathcal{T}$, the goal is to generate an augmented knowledge graph and jointly learn $d$ ($d<<|\mathcal{E}|$) dimensional embeddings for knowledge graph entities, which preserve structural and semantic properties of the knowledge graph. The learned representations are then used for the tasks of link prediction and node classification. Link prediction is defined as a binary classification whose goal is to predict whether or not an edge exists in KG, and node classification is the task of determining node labels in labelled graphs.

To address the problem of knowledge graph enrichment and embedding, we propose \textsc{Edge}, a framework that contains two major components, \ie augmented knowledge graph construction, and knowledge graph alignment in a joint embedding space.


\begin{table*}[t]
    \centering
    \caption{We employ representation learning algorithms to find a set of semantically and structurally similar entities to each target entity (column 2). We then find a set of keywords, $K$, that are representative of the target entity (column 3) and use them to query an external text and obtain a set of sentences, $S$ (column 4). Finally, we extract textual entities (column 5), and connect them to the target entity.}
    \label{tab:gt_gen}
    \resizebox{\textwidth}{!}{%
    \begin{tabular}{|l|m{1em}|l|l|l|l|} \hline
    \multicolumn{2}{|l|}{Target Entity} & semantically and structurally Similar Entities & \begin{tabular}[c]{@{}l@{}}Most definitive \\ keywords\end{tabular} & Sentences obtained from auxiliary text & \begin{tabular}[c]{@{}l@{}}Entities obtained from\\ information extraction\end{tabular} \\ \hline
    \multirow{2}{*}{\begin{tabular}[c]{@{}l@{}}Nonvenomous insect bite\\ of hip without infection\end{tabular}} & \rotatebox{90}{semantic}& \begin{tabular}[c]{@{}l@{}}1. Nonvenomous insect bite of foot with infection\\ 2. Crushing injury of hip and/or thigh\\ 3. Superficial injury of lip with infection\\ 4. Infected insect bite of hand\end{tabular} & \multirow{2}{*}{\begin{tabular}[c]{@{}l@{}}1. bite\\ 2. insect\\ 3. nonvenomous\\ 4. infect\end{tabular}} & \multirow{2}{*}{\begin{tabular}[c]{@{}l@{}}1. a wound resulting from biting by an animal or a person\\ 2. small air-breathing arthropod\\ 3. not producing or resulting from poison\\ 4. contaminate with a disease or microorganism\end{tabular}} & \multirow{2}{*}{\begin{tabular}[c]{@{}l@{}}1. wound\\ 2. arthropod\\ 3. poison\\ 4. microorganism\end{tabular}} \\ \cline{2-3}
     & \rotatebox{90}{structural} & \begin{tabular}[c]{@{}l@{}}1. Insect bite, nonvenomous, of back\\ 2. Tick bite\\ 3. Animal bite of calf\\ 4. Inset bite, nonvenomous, of foot and toe\end{tabular} &  &  &  \\ \hline
    \multirow{2}{*}{\begin{tabular}[c]{@{}l@{}}Insect bite, nonvenomous,\\ of foot and toe infected\end{tabular}} & \rotatebox{90}{semantic} & \begin{tabular}[c]{@{}l@{}}1. Insect bite, nonvenomous, of lower limb, infected\\ 2. Infected insect bite of hand\\ 3. Insect bite, nonvenomous, of hip\\ 4. Insect bite granuloma\end{tabular} & \multirow{2}{*}{\begin{tabular}[c]{@{}l@{}}1. bite\\ 2. insect\\ 3. lower\\ 4. skin\end{tabular}} & \multirow{2}{*}{\begin{tabular}[c]{@{}l@{}}1. a wound resulting from biting by an animal or a person\\ 2. small air-breathing arthropod\\ 3. move something or somebody to a lower position\\ 4. a natural protective body\end{tabular}} & \multirow{2}{*}{\begin{tabular}[c]{@{}l@{}}1. wound\\ 2. arthropod\\ 3. position\\ 4. protective body\end{tabular}} \\ \cline{2-3}
     & \rotatebox{90}{structural} & \begin{tabular}[c]{@{}l@{}}1. Nonvenomous insect bite of hip without infection\\ 2. Insect bite, nonvenomous, of back\\ 3. Recurrent infection of skin\\ 4. Skin structure of lower leg\end{tabular} &  &  &  \\ \hline
    \end{tabular}%
    }
\end{table*}

\subsection{Augmented Knowledge Graph Construction}
\label{subsec:GT}

Given the entities of $\mathcal{KG}$ and an external source of textual data, $\mathcal{T}$, we aim to generate an augmented graph, $a\mathcal{KG}$, which is a supergraph of $\mathcal{KG}$ (\ie $\mathcal{KG}$ is a subgraph of $a\mathcal{KG}$). Augmentation is the process of adding new entities to $\mathcal{KG}$. These newly added entities are called \emph{textual entities} or \emph{textual nodes}. A crucial property of $a\mathcal{KG}$ is that it contains entities of $\mathcal{KG}$. The presence of these entities establishes a relationship between the two graphs, and such a relationship will be leveraged to learn the shared graph embeddings. To construct $a\mathcal{KG}$, we need to find a set of keywords to query an external source, To obtain high quality keywords and acquire new textual entities, we design the following procedure per target entity $e_t$ (For every step of this process refer to Table \
\ref{tab:gt_gen} for a real example from SNOMED dataset). 

First, we find a set of semantically and structurally similar entities to $e_t$ denoted by $\mathcal{E}_{e_t}$. This set creates a textual context around $e_t$ which we use to find keywords to query an external text, \eg WordNet or Wikipedia. Here by \textit{query} we mean using the API of the external text to find related sentences, $S$ (for instance for a given keyword ``bite'' we can capture several sentences from the wikipedia page for the entry ``biting'' or find several Synsets\footnote{Synset is the fundamental building block of WordNet which is accompanied by a definition, example(s), etc.} from WordNet when we search for ``bite'').

Finally, we extract entities from $S$ and attach them to $e_t$. We call these new entities, \emph{textual entities} or \emph{textual features}. By connecting these newly found textual entities to the $e_t$, we enhance $\mathcal{KG}$ and generate the augmented knowledge graph, $a\mathcal{KG}$. We observed that the new textual entities are different from our initial feature space. Also, it is possible that two different target entities share one or more textual nodes, hence the distance between them in $a\mathcal{KG}$ would decrease. The implementation details of this process is provided in Supplementary materials.

Querying an external text allows us to extend the feature space beyond the context around $e_t$. By finding other entities in $\mathcal{KG}$ that are similar to the target entity and extracting keywords from the collection of them to query the external text, distant entities that are related but not connected would become closer to each other owing to the shared keywords.

Figure~\ref{fig:gt} illustrates a subset of SNOMED graph and its augmented counterpart by following the above procedure. As this figure reveals, the structure of $a\mathcal{KG}$ is different from $\mathcal{KG}$, and as a result of added textual nodes, distant but similar entities would become closer. Therefore, augmenting knowledge graphs would alleviate the KG sparsity issue. Although we may introduce noise by adding new entities but later in the alignment process we address this issue.

\emph{\textbf{Remarks.}} In the above procedure, we need to obtain similar entities before looking for textual entities, and the rationality of such a strategy is discussed as follows. One naive approach is to simply use keywords included in the target entity to find new textual features. In this way, we would end up with textual features that are related to that target entity, but we cannot extend the feature space to capture similarity (\ie dependency) among entities.

\subsection{Knowledge Graph Alignment in Joint Embedding Space}

With the help of augmented knowledge graph $a\mathcal{KG}$, we aim to enrich the graph embeddings of $\mathcal{KG}$. However, inevitably, a portion of newly added entities are noisy, and even potentially wrong. To mitigate this issue, we are inspired by Hinton et al.~\cite{hinton2015distilling}, and propose a graph alignment process for knowledge distillation. In fact, $a\mathcal{KG}$ and $\mathcal{KG}$ share some common entities, which makes it possible to map two knowledge graphs into a joint embedding space. In particular, we propose to extract low-dimensional node embeddings of two knowledge graphs using graph auto-encoders~\cite{kipf2016variational}, and design novel constraints to align two graphs in the embedding space. The architecture of our approach is illustrated in Figure~\ref{fig:model}.

Let $\textbf{A}_K$ and $\textbf{A}_T$ denote the adjacency matrices of $\mathcal{KG}$ and $a\mathcal{KG}$, respectively. The loss functions of graph auto-encoders that reconstruct knowledge graphs are defined as: 
\begin{equation}
    \mathcal{L}_K=\min_{\textbf{Z}_K}||\textbf{A}_K-\hat{\textbf{A}}_K||_2,
    \label{eq:lk}
\end{equation}
\begin{equation}
    \mathcal{L}_T=\min_{\textbf{Z}_T}||\textbf{A}_T-\hat{\textbf{A}}_T||_2,
    \label{eq:lt}
\end{equation}
where $\hat{\textbf{A}}_K = \sigma(\textbf{Z}_K\textbf{Z}_K^{\top})$ is the reconstructed graph using node embeddings $\textbf{Z}_K$. And $\textbf{Z}_K$ is the output of graph encoder that is implemented by a two-layer GCN \cite{kipf2016variational}:
\begin{equation}
\textbf{Z}_K = \text{GCN}(\textbf{A}_K,\textbf{X}_K)=\tilde{\textbf{A}_K}\ \tanh(\tilde{\textbf{A}}_K\textbf{X}_K\textbf{W}_0)\textbf{W}_1,
\label{eq:gcn}
\end{equation}
where $\tilde{\textbf{A}}_K=\textbf{D}_K^{-\frac{1}{2}}\textbf{A}_K\textbf{D}_K^{-\frac{1}{2}}$. $\textbf{D}_K$ is the degree matrix, $\tanh{(.)}$ is the Hyperbolic Tangent function that acts as the activation function of the neurons, $\textbf{W}_i$ are the model parameters, and $\textbf{X}_K$ is the feature matrix.\footnote{In case of a featureless graph, an identity matrix, $\textbf{I}$, replaces $\textbf{X}_K$.} Similarly, $\hat{\textbf{A}}_T = \sigma(\textbf{Z}_T\textbf{Z}_T^{\top})$, and $\textbf{Z}_T$ is learned by another two-layer GCN. Equations~\eqref{eq:lk} and~\eqref{eq:lt} are \textit{$l_2$-norm} based loss functions that aim to minimize the distance between original graphs and the reconstructed graphs. 

Furthermore, to map $\mathcal{KG}$ and $a\mathcal{KG}$ to a joint embedding space and align their embeddings through common entities, we define the following graph alignment loss function:
\begin{equation}
    \mathcal{L}_J=||\mathbf{Z}_K-\mathbf{R}\mathbf{Z}_T||_2,
    \label{eq:lj}
\end{equation}
where $\mathbf{R}$ is a transform matrix that selects common entities that exist in $\mathcal{KG}$ and $a\mathcal{KG}$. Note that the two terms $\mathbf{Z}_K$ and $\mathbf{R}\mathbf{Z}_T$ should be of the same size in the $L_2$ norm equation. Our motivation is to align the embeddings of common entities across two knowledge graphs. By using $\mathbf{R}$, the node embeddings of common entities can be selected from $\mathbf{Z}_T$. Note that $\mathbf{Z}_T$ is always larger than $\mathbf{Z}_K$, as $\mathcal{KG}$ is a subgraph of $a\mathcal{KG}$. Equation~\eqref{eq:lj} also helps preserve local structures of the original knowledge graph $\mathcal{KG}$ in the graph embedding space. In other words, nodes that are close to each other in the original knowledge graph will be neighbors in the augmented graph as well.

\begin{algorithm}[t]
\caption{Training process of \textsc{Edge}}
\begin{algorithmic}[1]
\small
    \Require $\textbf{A}_K$, $\textbf{X}_K$, $\textbf{A}_T$, $\textbf{X}_T$, $\text{POS}$, $\text{NEG}$, 
    \Require $\textbf{R}\in \mathbb{R}^{|\mathcal{E}_K|\times(|\mathcal{E}_T|-|\mathcal{E}_K|)}$
    \For{each epoch}
        \State $\hat{\textbf{A}}_K=\sigma(\textbf{Z}_K\textbf{Z}_K^{\top})$
        \State $\textbf{Z}_K=\tilde{\textbf{A}}_K\tanh(\tilde{\textbf{A}}_K\textbf{X}_K\textbf{W}_0^K)\textbf{W}_1^K$ 
        \State $\hat{\textbf{A}}_T=\sigma(\textbf{Z}_T\textbf{Z}_T^{\top})$
        \State $\textbf{Z}_T=\tilde{\textbf{A}}_T\tanh(\tilde{\textbf{A}}_T\textbf{X}_T\textbf{W}_0^T)\textbf{W}_1^T$ 
        \State Calculate $\mathcal{L}_K$ and $\mathcal{L}_T$  using Equations \eqref{eq:lk} and \eqref{eq:lt}.
        \State Compute $\mathcal{L}_J$ using Equation \eqref{eq:lj}
        \State Find negative and positive samples and calculate $\mathcal{L}_N$ using Equation \eqref{eq:ln}
        \State Sum up all losses with their corresponding ratios using Equation \eqref{eq:loss}
        \State Run Adam optimizer to minimize $\mathcal{L}$
        \State Update model parameters $\textbf{W}_i^K$ and $\textbf{W}_i^T$
    \EndFor
    \Ensure $Z_K$
\end{algorithmic}
\label{alg:loss}
\end{algorithm}

Moreover, we notice that the proposed augmented knowledge graph $a\mathcal{KG}$ involves more complicated structures than the original knowledge graph $\mathcal{KG}$, due to the newly added textual nodes for each target entity in $\mathcal{KG}$. In $a\mathcal{KG}$, one target entity is closely connected to its textual nodes, and their embeddings should be very close to each other in the graph embedding space. However, such local structures might be distorted in the graph embedding space. Without proper constraints, it is possible that one target entity is close to textual entities of other target entities in the embedding space, which is undesired for downstream applications. To address this issue, we design a margin-based loss function with negative sampling to preserve the locality relationship as follows:
\begin{equation}
    \mathcal{L}_N=-\log(\sigma(\textbf{z}_e^{\top}\textbf{z}_t))-\log(\sigma(-\textbf{z}_e^{\top}\textbf{z}_{t'})),
    \label{eq:ln}
\end{equation}
where $\textbf{z}_t$ are the embeddings of the related textual nodes, $\textbf{z}_t'$ are the embeddings of textual nodes that are not related to the target entity, and $\sigma$ is the \textit{sigmoid} function.

Finally, the overall loss function is defined as:
\begin{equation}
    \mathcal{L}=\min_{\textbf{Z}_K, \textbf{Z}_T} \underbrace{\mathcal{L}_K+\alpha\mathcal{L}_T}_{\substack{\text{reconstruction}\\ \text{loss}}}+\underbrace{\beta\mathcal{L}_J}_{\substack{\text{graph}\\ \text{alignment}}}+\underbrace{\gamma\mathcal{L}_N}_{\substack{\text{locality}\\ \text{preserving}}},
    \label{eq:loss}
\end{equation}
where $\alpha$, $\beta$, and $\gamma$ are hyper-parameters. We perform full-batch gradient descent using the Adam optimizer to learn all the model parameters in an end-to-end fashion. The whole training process of our approach is summarized in Algorithm \ref{alg:loss}.

The learned low-dimensional node embeddings $\textbf{Z}_K$ could benefit a number of unsupervised and supervised downstream applications, such as link prediction and node classification. Link prediction is the task of inferring missing links in a graph, and node classification is the task of predicting labels to vertices of a (partially) labeled graph. Extensive evaluations on both tasks will be provided in the experiment section.

\begin{table*}[t]
\centering
\caption{Link prediction results for SNOMED and three citation networks. Numbers for SNOMED are obtained from rerunning their code on the dataset. The rest of the results are reported from corresponding papers.}
\resizebox{.80\textwidth}{!}{
\begin{tabular}{@{}ccccccccc@{}}
\toprule
\multirow{2}{*}{Model} & \multicolumn{2}{c}{SNOMED} & \multicolumn{2}{c}{Cora} & \multicolumn{2}{c}{Citeseer} & \multicolumn{2}{c}{PubMed} \\ \cmidrule(l){2-9} 
 & AUC & \multicolumn{1}{c|}{AP} & AUC & \multicolumn{1}{c|}{AP} & AUC & \multicolumn{1}{c|}{AP} & AUC & AP \\ \midrule
\multicolumn{1}{c|}{GAE \cite{kipf2016variational}} & 0.773 & \multicolumn{1}{c|}{0.844} & 0.914 & \multicolumn{1}{c|}{0.926} & 0.908 & \multicolumn{1}{c|}{0.920} & 0.964 & 0.965 \\
\multicolumn{1}{c|}{LoNGAE \cite{tran2018learning}} & 0.890 & \multicolumn{1}{c|}{0.910} & 0.954 & \multicolumn{1}{c|}{0.963} & 0.953 & \multicolumn{1}{c|}{0.961} & 0.960 & 0.963 \\
\multicolumn{1}{c|}{ARVGE \cite{pan2018adversarially}} & 0.805 & \multicolumn{1}{c|}{0.864} & 0.924 & \multicolumn{1}{c|}{0.926} & 0.924 & \multicolumn{1}{c|}{0.930} & 0.968 & 0.971 \\
\multicolumn{1}{c|}{SCAT \cite{zou2019encoding}} & 0.902 & \multicolumn{1}{c|}{0.918} & 0.945 & \multicolumn{1}{c|}{0.946} & 0.973 & \multicolumn{1}{c|}{\textbf{0.976}} & \textbf{0.975} & \textbf{0.972} \\
\multicolumn{1}{c|}{GIC \cite{mavromatis2020graph}} & - & \multicolumn{1}{c|}{-} & 0.935 & \multicolumn{1}{c|}{0.933} & 0.970 & \multicolumn{1}{c|}{0.968} & 0.937 & 0.935 \\\midrule
\multicolumn{1}{c|}{\textsc{Edge} (This work)} & \textbf{0.916} & \multicolumn{1}{c|}{\textbf{0.944}} & \textbf{0.973} & \multicolumn{1}{c|}{\textbf{0.975}} & \textbf{0.974} & \multicolumn{1}{c|}{\textbf{0.976}} & 0.969 & 0.968 \\ \bottomrule
\end{tabular}
\label{tab:lp}
}
\vspace{-4mm}
\end{table*}

\subsection{Model Discussions}
We have proposed a general framework for graph enrichment and embedding by exploiting auxiliary knowledge sources. What we consider as a source of knowledge is a textual knowledge base that can provide additional information about the entities of the original knowledge graph. It is a secondary source of knowledge that supplies new sets of features outside of the existing feature space, which improves the quality of representations.

The proposed graph alignment approach can fully exploit augmented knowledge graph and thus improve the graph embeddings. Although $a\mathcal{KG}$ is a supergraph of $\mathcal{KG}$, its connectivity pattern is different. With the help of our customized loss function for graph alignment, both graphs contribute in the quality of derived embeddings. We will also demonstrate the superiority of our joint embedding approach over the independent graph embedding approach (with only $a\mathcal{KG}$) in the experiments, and we investigate which component of our model contributes more in the final performance in the ablation study in Subsection \ref{subsec:abl}.

\section{Experiment}
\label{sec:experiment}
We design our experiments to investigate effectiveness of different components of \textsc{Edge} as well as its overall performance. To this end, we aim to answer the following three questions\footnote{We plan to release our code upon publication.}.
\begin{enumerate}[label=\subscript{Q}{{\arabic*}}]
\setlength\itemsep{0em}
    \item How well does \textsc{Edge} perform compared to state-of-the-art in the task of link prediction? (Section \ref{subsec:lp})
    \item How is the quality of embeddings generated by \textsc{Edge} compared to similar methods? (Sections \ref{subsec:nc} and \ref{subsec:ee})
    \item What is the contribution of each component (augmentation and alignment) in the overall performance? (Section \ref{subsec:abl})
\end{enumerate}

\subsection{Task 1: Link Prediction}
\label{subsec:lp}
To investigate Q1 we perform link prediction on four benchmark datasets, and compare the performance of our model with five relevant baselines. For this task we consider SNOMED and three citation networks. For SNOMED, similar to \cite{kartsaklis2018mapping}, we select 21K medical concepts from the original dataset. Each entity in SNOMED is a text description of a medical concept, \eg \textit{Nonvenomous insect bite of hip without infection}. According to the procedure explained in subsection \ref{subsec:GT}, we construct an augmented knowledge graph, $a\mathcal{KG}$. Additionally, we consider three other datasets, namely Cora, Citeseer, and PubMed, which are citation networks consisting of 2,708, 3,312, and 19,717 papers, respectively. In all three datasets, a short text accompanies each node which is extracted from the title or abstract of the paper. For these networks, \textit{relation} is defined as citation and the textual content of the nodes enables us to obtain $a\mathcal{KG}$. Cora and Citeseer datasets come with a set of default features. We defer the detailed description of datasets in the supplementary. 

In this experiment, for each dataset, we train the model on 85\% of the input graph. Other 15\% of the data is split into 5\% validation set and 10\% as part of the test set (positive samples only). An additional set of edges are produced, equal to the number of positive samples, which does not exist in the graph, as negative samples. The union of positive and negative samples are used as the test set. In all baselines, we test the model on $\mathcal{KG}$. We obtain the following values for loss ratios after hyper-parameter tuning: $\alpha=0.001$, $\beta=10$, $\gamma=1$. We discuss parameter tuning and explain the small value of $\alpha$ in Section \ref{subsec:param}.

We provide comparison against VGAE \cite{kipf2016variational} and its adversarial variant ARVGE \cite{pan2018adversarially}. Also we consider LoNGAE \cite{tran2018learning}, SCAT \cite{zou2019encoding} and GIC \cite{mavromatis2020graph} which are designed for link prediction task on graphs, hence they make strong baselines. Table \ref{tab:lp} presents the Area Under the ROC Curve (AUC) and average precision (AP) scores for five baselines and our methods across all datasets. We observe that \textsc{Edge} outperforms all baselines in three out of four datasets and produces comparable results for PubMed dataset.

\begin{table}[t]
\centering
\caption{Node classification results in terms of accuracy for citation networks. TR stands for training ratio and \textit{un.} and \textit{semi.} are short for unsupervised and semi-supervised.}
\label{tab:node_clf}
\resizebox{\columnwidth}{!}{%
\begin{tabular}{ccccc}
\hline
Model & Approach & \begin{tabular}[c]{@{}c@{}}Cora \\ TR=0.5\end{tabular} & \begin{tabular}[c]{@{}c@{}}Citeseer\\ TR=0.03\end{tabular} & \begin{tabular}[c]{@{}c@{}}PubMed\\ TR=0.003\end{tabular} \\ \hline
\multicolumn{1}{c|}{DeepWalk} & \textit{un.} & 0.67 & 0.43 & 0.65 \\
\multicolumn{1}{c|}{GCN} & \textit{semi.} & 0.81 & 0.70 & 0.79 \\
\multicolumn{1}{c|}{GAT} & \textit{semi.} & \textbf{0.83} & \textbf{0.72} & 0.79 \\
\multicolumn{1}{c|}{LoNGAE} & \textit{semi.} & 0.78 & 0.71 & 0.79 \\
\multicolumn{1}{c|}{MixHop} & \textit{semi.} & 0.82 & 0.71 & \textbf{0.81} \\ \hline
\multicolumn{1}{c|}{\textsc{Edge}} & \textit{un.} & 0.81 & 0.66 & 0.76 \\ \hline
\end{tabular}%
}
\vspace{-4mm}
\end{table}

\subsection{Task 2: Node Classification on Citation Networks}
\label{subsec:nc}
To evaluate the quality of embeddings (Q2) we design a node classification task based on the final product of our model. For this task, we use Cora, Citeseer and PubMed datasets, and follow the same procedure explained in \ref{subsec:GT} to generate $a\mathcal{KG}$ and jointly map the two graphs into an embedding space. All the settings are identical to Task 1. To perform node classification, we use the final product of our model, which is a $160$ dimensional vector per node. We train a linear SVM classifier and obtain the accuracy measure to compare the performance of our model with state-of-the-art methods. Training ratio varies across different datasets, and we consider several baselines to compare our results against.

We compare our approach with state-of-the-art semi-supervised models for node classification, including GCN \cite{kipf2017semi}, GAT \cite{velickovic2018graph}, LoNGAE \cite{tran2018learning}, and MixHop \cite{Haija2019MixHop}. These models are semi-supervised, thus they were exposed to node labels during training while our approach is completely unsupervised. We also include DeepWalk, an unsupervised approach, to have a more complete view for our comparison. Table \ref{tab:node_clf} reveals that our model achieves reasonable performance compared with semi-supervised models in two out of three datasets. Since \textsc{Edge} is fully unsupervised, it is fair to declare that its performance is comparable as other methods are exposed to more information (\ie node labels).

\subsection{Embedding Effectiveness}
\label{subsec:ee}
\begin{figure}[t]
    \centering
    \begin{subfigure}{0.49\columnwidth}
        \centering
        \includegraphics[width=\textwidth]{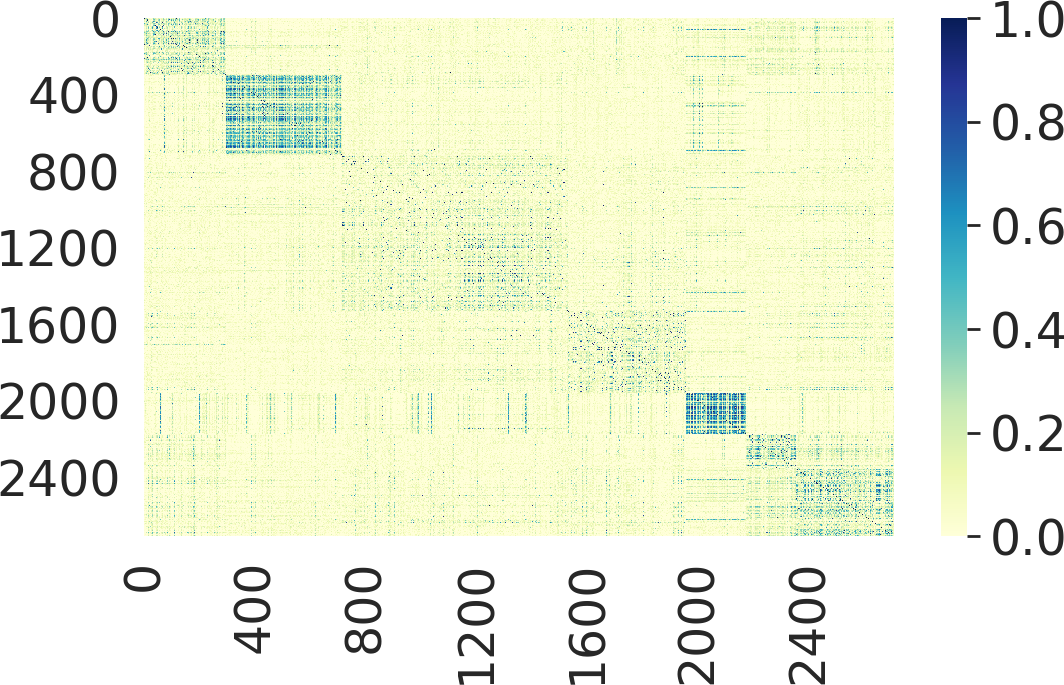}
        \caption{\textsc{Edge}}
        \label{subfig:EDGE_sim}
    \end{subfigure}
    \hfill
    \begin{subfigure}{0.49\columnwidth}
        \centering
        \includegraphics[width=\textwidth]{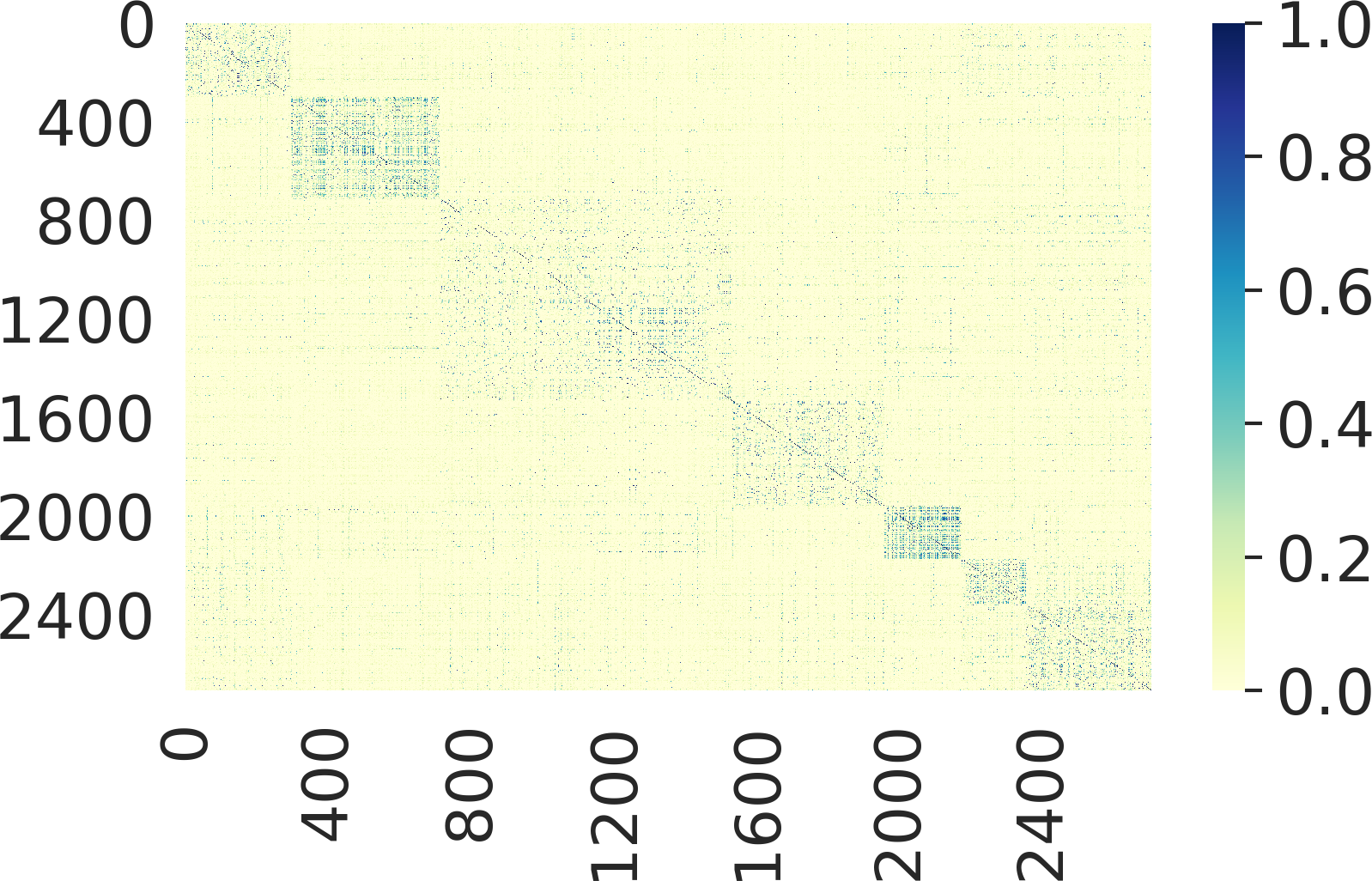}
        \caption{GAE}
        \label{subfig:gae_sim}
    \end{subfigure}
    \caption{Pair-wise similarity comparison between GAE and \textsc{Edge}.}
    \label{fig:sim}
\end{figure}

Further, to measure the quality of embeddings produced by our model and compare it against the baseline, we visualize the similarity matrix of node embeddings for two scenarios on the Cora dataset: 1) GAE on $\mathcal{KG}$, and 2) \textsc{Edge} on $\mathcal{KG}$ and $a\mathcal{KG}$. The results are illustrated in Figure \ref{fig:sim}. In this heatmap, elements are pair-wise similarity values sorted by different labels (7 classes). We can observe that the block-diagonal structure learned by our approach is clearer than that of GAE, indicating enhanced separability between different classes.

\begin{figure}
    \centering
    \begin{subfigure}{0.49\columnwidth}
        \centering
        \includegraphics[width=\textwidth]{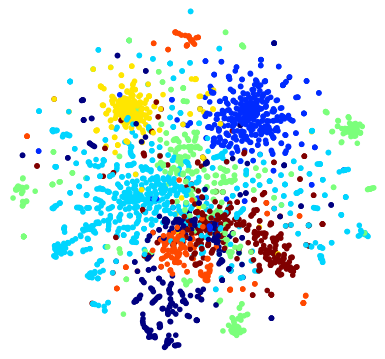}
        \caption{feature-based}
    \end{subfigure}
    \hfill
    \begin{subfigure}{0.49\columnwidth}
        \centering
        \includegraphics[width=\textwidth]{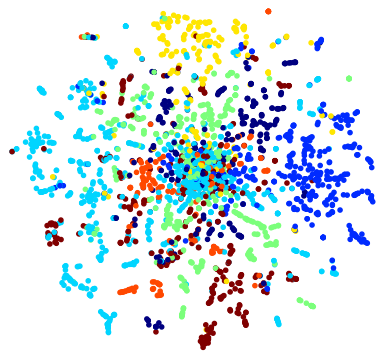}
        \caption{featureless}
    \end{subfigure}
    \caption{Visualization of embedding vectors on Cora: a) with and b) without features to study the effect of features on quality of embeddings in node classification.}
    \label{fig:tsne}
    \vspace{-2mm}
\end{figure}

Next, we examine our model in more details and study how different parameters affect its performance.

\begin{figure*}[t]
\vspace{-2mm}
    \centering
    \begin{subfigure}{0.24\textwidth}
        \centering
        \includegraphics[width=\textwidth]{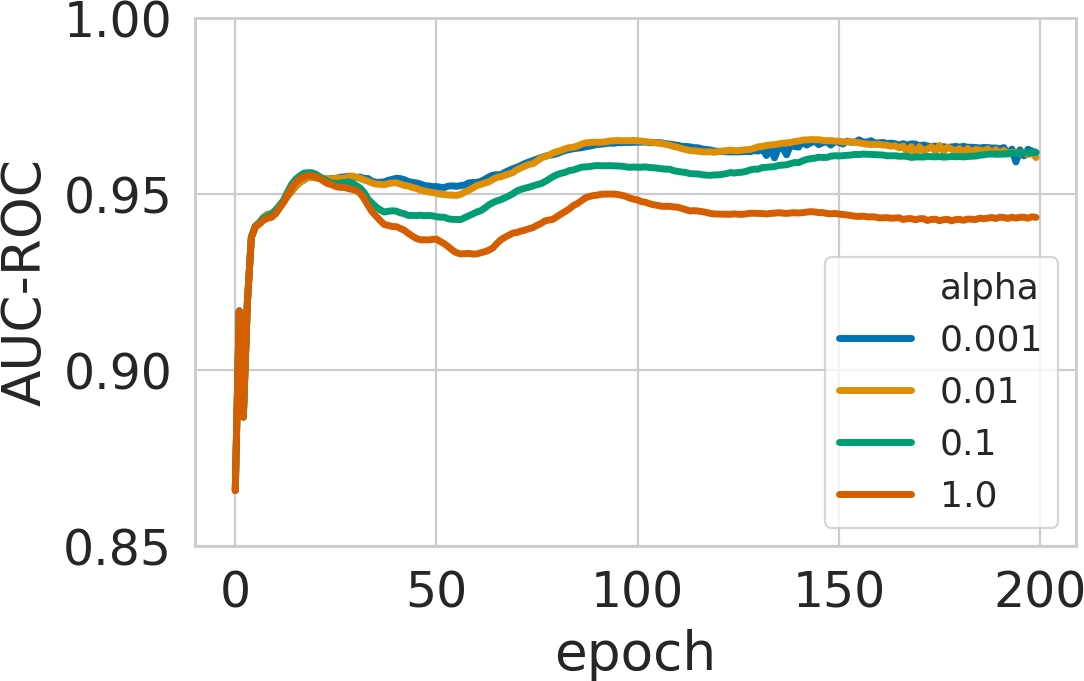}
        \caption{Cora $\beta=1$, $\gamma=1$}
        \label{subfig:cora_beta1gamma1}
    \end{subfigure}
    \hfill
    \begin{subfigure}{0.24\textwidth}
        \centering
        \includegraphics[width=\textwidth]{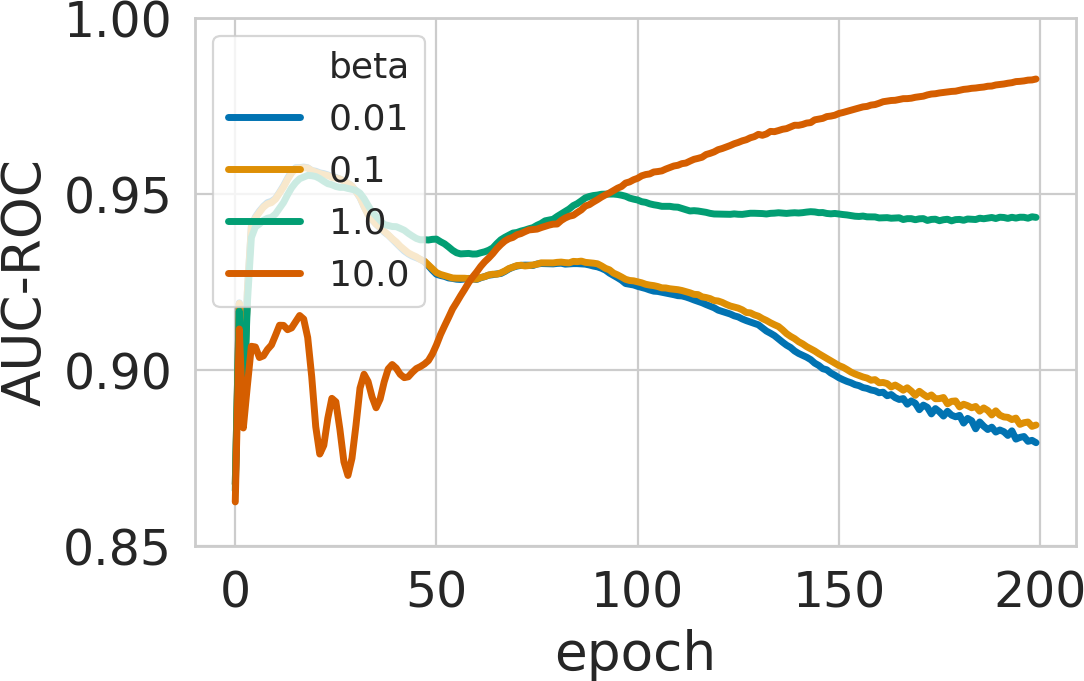}
        \caption{Cora $\alpha=1$, $\gamma=1$}
        \label{subfig:cora_alpha1gamma1}
    \end{subfigure}
    \hfill
    \begin{subfigure}{0.24\textwidth}
        \centering
        \includegraphics[width=\textwidth]{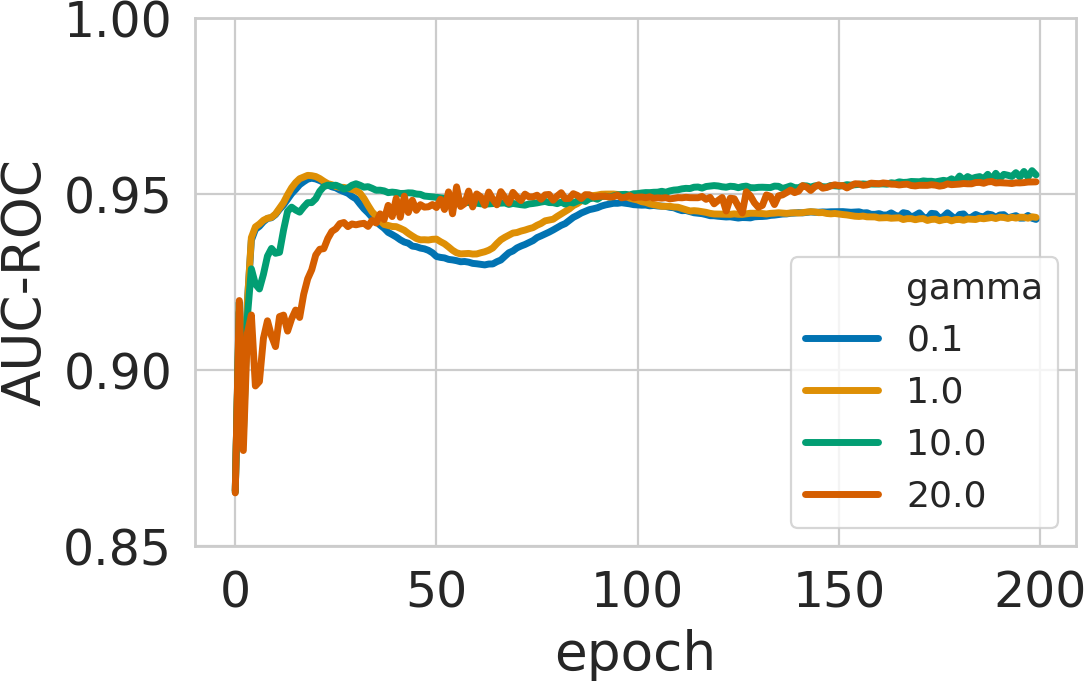}
        \caption{Cora $\alpha=1$, $\beta=1$}
        \label{subfig:cora_alpha1beta1}
    \end{subfigure}
    \hfill
    \begin{subfigure}{0.24\textwidth}
        \centering
        \includegraphics[width=\textwidth]{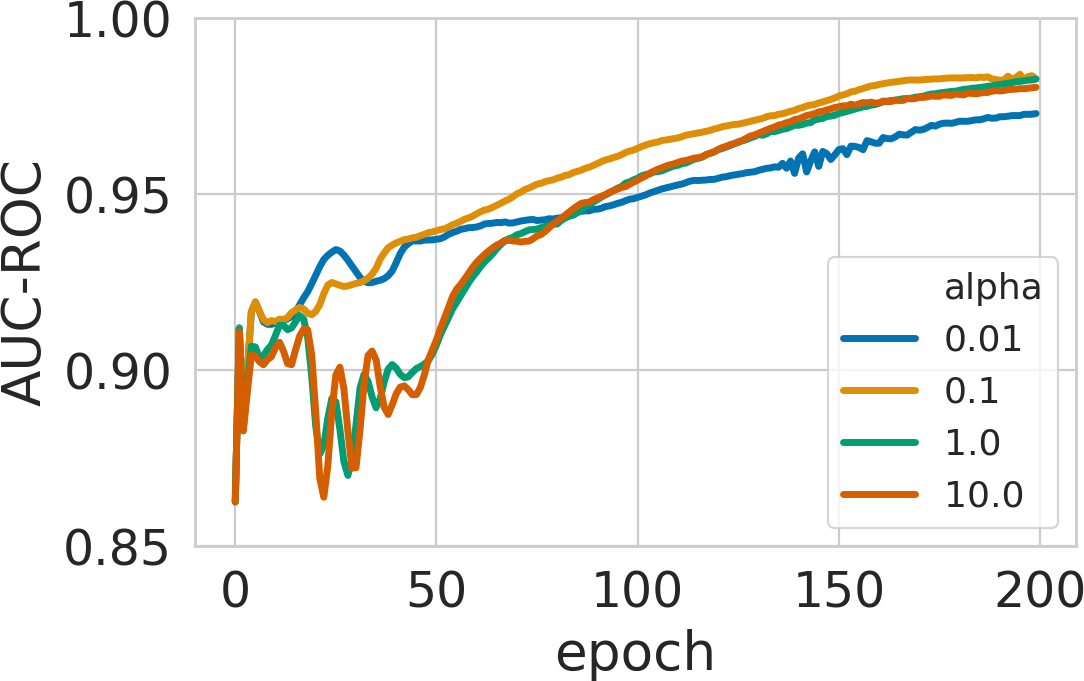}
        \caption{Cora $\beta=10$, $\gamma=1$}
        \label{subfig:cora_beta10gamma1}
    \end{subfigure}
    \vspace{-2mm}
    \caption{Effect of parameterization on link prediction performance}
    \vspace{-2mm}
    \label{fig:param}
\end{figure*}

\subsection{Ablation Study}
\label{subsec:abl}
To investigate the effectiveness of different modules of our model (Q3), we consider two scenarios. First we use a single graph to train our model. Note that when we use a single graph, the graph alignment and locality preserving losses are discarded and our model is reduced to GAE. In single graph scenario we consider two versions of augmented graph, $a\mathcal{KG}$ that was explained in subsection \ref{subsec:GT} and $a\mathcal{KG}^*$ that was created based on co-occurrence proposed by \cite{kartsaklis2018mapping}. In the second scenario, we use two graphs to jointly train \textsc{Edge}, and we feed our model with $\mathcal{KG}+a\mathcal{KG}^*$ and $\mathcal{KG}+a\mathcal{KG}$ to show the effect of augmentation.

\begin{table}[t]
	\centering
	\caption{Link prediction results for SNOMED dataset. In this table $a\mathcal{KG}^*$ is the augmented knowledge graph generated using the method explained in \cite{kartsaklis2018mapping}}
	\label{tab:lp-snomed}
	\resizebox{\columnwidth}{!}{%
	\begin{tabular}{@{}llcc@{}}
		\toprule
		Input & Model & AUC & AP \\ \midrule
		$\mathcal{KG}$ & GAE~\citep{kipf2016variational}  & 0.77 & 0.84 \\
		$a\mathcal{KG}^*$ & GAE \cite{kipf2016variational} &  0.85 & 0.88 \\
		$a\mathcal{KG}$ & GAE \cite{kipf2016variational} & 0.86 & 0.90 \\
		$\mathcal{KG}+a\mathcal{KG}^*$ & \textsc{Edge} (This work) & 0.90 & 0.93 \\
		$\mathcal{KG}+a\mathcal{KG}$ & \textsc{Edge} (This work)& \textbf{0.91} & \textbf{0.94} \\ 
		\bottomrule
	\end{tabular}
    }
    \vspace{-4mm}
\end{table}

For link prediction we only consider SNOMED dataset which is the largest dataset, and as Table \ref{tab:lp-snomed} presents we observe that our augmentation process is slightly more effective than co-occurrence based augmentation. More importantly, by comparing second two rows with first two rows we realize that alignment module improves the performance more than augmentation process which highlights the importance of our proposed joint learning method. Moreover, we repeat this exercise for node classification (see Table \ref{tab:node_clf_abl}) which results in a similar trend across all datasets.

Finally, we plot the t-SNE visualization of embedding vectors of our model with and without features. Figure \ref{fig:tsne} clearly illustrates the distinction between quality of the clusters for the two approaches. This implies that knowledge graph text carries useful information. When the text is incorporated into the model, it can help improve the model performance. 

\subsection{Parameter Sensitivity}
\label{subsec:param}
We evaluate the parameterization of \textsc{Edge}, and specifically we examine how changes to hyper parameters of our loss function (\ie $\alpha$, $\beta$ and $\gamma$) could affect the model performance in the task of link prediction on Cora dataset. In each analysis, we fix the values of two out of three parameters and study the effect of the variation of the third parameter on evaluating AUC scores across 200 epochs. The detailed results are shown in Figure \ref{fig:param}.

Figure \ref{subfig:cora_beta1gamma1} shows the effect of varying $\alpha$, when $\beta=1$ and $\gamma=1$ are fixed. We observe a somewhat consistent trend across performance for different values of $\alpha$. It is evident that decreasing $\alpha$ improves the performance. $\alpha$ is the coefficient of $\mathcal{L}_T$ (see Equation \ref{eq:lt}). This examination suggests that the effect of this loss function is less significant, because we re-address it in the $\mathcal{L}_N$ part of the loss function, where we consider the same graph ($a\mathcal{KG}$) and try to optimize distance between its nodes but with more constraints.

Figure \ref{subfig:cora_alpha1gamma1} illustrates the effect of varying $\beta$, while $\alpha=1$ and $\gamma=1$ are fixed. Tuning $\beta$ results in more radical changes in the model performance, which is again consistent between the two datasets. Small values for $\beta$ degrades performance remarkably, and we observe a much more improved AUC score for larger values of $\beta$. This implies the dominant effect of the joint loss function, $\mathcal{L}_J$, which is defined as the distance between corresponding entities of $\mathcal{KG}$ and $a\mathcal{KG}$.

\begin{table}[t]
\centering
\caption{Node classification results in terms of accuracy for citation networks. TR stands for training ratio and $a\mathcal{KG}^*$ is an augmented knowledge graph produced by the method proposed in \cite{kartsaklis2018mapping}}
\label{tab:node_clf_abl}
\resizebox{\columnwidth}{!}{%
\begin{tabular}{llccc}
\hline
Input & Model & \begin{tabular}[c]{@{}c@{}}Cora \\ TR=0.5\end{tabular} & \begin{tabular}[c]{@{}c@{}}Citeseer\\ TR=0.03\end{tabular} & \begin{tabular}[c]{@{}c@{}}PubMed\\ TR=0.003\end{tabular} \\ \hline
\multicolumn{1}{l}{$\mathcal{KG}$} & GAE & 0.62 & 0.51 & 0.60 \\
\multicolumn{1}{l}{$a\mathcal{KG}^*$} & GAE & 0.70 & 0.57 & 0.65 \\
\multicolumn{1}{l}{$a\mathcal{KG}$} & GAE & 0.75 & 0.61 & 0.67 \\
\multicolumn{1}{l}{$\mathcal{KG}+a\mathcal{KG}^*$} & \textsc{Edge} & 0.80 & 0.64 & 0.73 \\
\multicolumn{1}{l}{$\mathcal{KG}+a\mathcal{KG}$} & \textsc{Edge} & \textbf{0.81} & \textbf{0.66} & \textbf{0.76} \\ \hline
\end{tabular}%
}
\vspace{-4mm}
\end{table}

Next, we fix $\alpha=1$ and $\beta=1$ and tweak $\gamma$ from $0.1$ to $10$. As Figure \ref{subfig:cora_alpha1beta1} reveals, the variation in performance is very small. Finally, as we obtained the best results when $\beta=10$, we set $\gamma=1$ and once again tune $\alpha$. Figure \ref{subfig:cora_beta10gamma1} shows the results for this updated setting. These experiments confirm the insignificance of parameter $\alpha$. In practice, we obtained the best results by setting $\alpha$ to 0.001.


\section{Conclusion}
\label{sec:con}
Sparsity is a major challenge in KG embedding, and many studies failed to properly address this issue. We proposed \textsc{Edge}, a novel framework to enrich KG and align the enriched version with the original one with the help of auxiliary text. Using external source of information introduces new sets of features that enhance the quality of embeddings. We applied our model on three citation networks and one large scale medical knowledge graph. Experimental results show that our approach outperforms existing graph embedding methods on link prediction and node classification. 


\section*{Acknowledgment}
This research is supported in part by the U.S. Army Research Office Award under Grant Number W911NF-21-1-0109, and a gift funding from Adobe Research. 

\balance
\bibliography{main.bib}
\bibliographystyle{acl_natbib}

\newpage
\appendix


\end{document}